\documentclass[11pt, oneside]{article}
\usepackage[utf8]{inputenc}
\usepackage[margin=1in]{geometry}
\usepackage{titlesec}
\usepackage{authblk}
\usepackage{adjustbox}
\usepackage[font=small,skip=5pt]{caption}
\usepackage{amssymb, amsmath, amsthm}
  \DeclareMathOperator*{\argmax}{argmax}
  
\usepackage{mathtools}
\usepackage{bm}
\usepackage{todonotes}
\usepackage{subcaption}
\usepackage{graphicx}
  \graphicspath{ {images/} }
\usepackage{float}
\usepackage{url}
\usepackage{hyperref}
\hypersetup{
  citecolor=blue,
  linkcolor=blue,
  colorlinks=true,
  urlcolor=blue
}
\usepackage[backend=biber, style=numeric-comp, url=false, doi=false,
            isbn=false, eprint=false, natbib=true, maxbibnames=20,
            sorting=none, uniquelist=false]{biblatex}

\AtEveryBibitem{\clearfield{month}}
\AtEveryBibitem{\clearfield{day}}
\AtEveryBibitem{\clearfield{version}}

\title{The case for fully Bayesian optimisation in small-sample trials}

\author[]{Yuji Saikai}
\affil[]{School of Mathematics and Statistics, the University of Melbourne}

\date{}

\begin{document}
%%%%%%%%%%%%%%%%%%%%%%%%%%%%%%

\maketitle
\begin{abstract}
\noindent

While sample efficiency is the main motive for use of Bayesian optimisation when black-box functions are expensive to evaluate, the standard approach based on type II maximum likelihood (ML-II) may fail and result in disappointing performance in small-sample trials. The paper provides three compelling reasons to adopt fully Bayesian optimisation (FBO) as an alternative. First, failures of ML-II are more commonplace than implied by the existing studies using the contrived settings. Second, FBO is more robust than ML-II, and the price of robustness is almost trivial. Third, FBO has become simple to implement and fast enough to be practical. The paper supports the argument using relevant experiments, which reflect the current practice regarding models, algorithms, and software platforms. Since the benefits seem to outweigh the costs, researchers should consider adopting FBO for their applications so that they can guard against potential failures that end up wasting precious research resources.

\vspace{10pt}
\noindent \textit{Keywords}: BoTorch, expected improvement, fully Bayesian Gaussian process, Pyro, type II maximum likelihood
\end{abstract}

\vspace{10pt}
%%%%%%%%%%%%%%%%%%%%%%%%%%%%%%
\section{Introduction}
%%%%%%%%%%%%%%%%%%%%%%%%%%%%%%
Sample efficiency is the predominant motive for use of Bayesian optimisation (BO) when optimising black-box functions that are expensive to evaluate. Function evaluations are said to be expensive when each evaluation costs a significant amount of resources including time and money. Function evaluations take place typically in the form of either real-world trials or computer simulations. While the former case clearly can be expensive, because of the lack of analytic expressions for objective functions and/or constraints, the latter case can also be expensive. For example, researchers used BO to learn robot controllers with only a handful of trials \citep{chatzilygeroudis_survey_2020}. Consequently, in practical applications of BO, small-sample performance is of the greatest concern.

Gaussian process (GP) is the de facto standard surrogate model for BO, and GPs must be specified in terms of kernel hyperparameters before applying BO. Since hyperparameters suitable for particular problems are rarely known beforehand, these are estimated based on observations and sequentially updated throughout the trials, commonly using type II maximum likelihood (ML-II) \citep{jones_efficient_1998, mackay_comparison_1999, lalchand_approximate_2020}. In practice, estimating hyperparameters with small samples is a challenging task \citep{wang_theoretical_2014}. In particular, maximisation of the marginal likelihood is numerically difficult because a wide range of length-scales, both very short and intermediate values, are generally consistent with small data \citep{rasmussen_gaussian_2006}. Several studies provide detailed accounts of how BO fails in various situations \citep{jones_taxonomy_2001, forrester_global_2008,bull_convergence_2011,benassi_robust_2011}. Nonetheless, despite the challenge, this is precisely the kind of problems for which researchers use BO---locating inputs that yield good enough function values using only small samples.

To address the challenge, the fully Bayesian approach has been proposed in the context of GP regression \citep{ohagan_curve_1978,handcock_bayesian_1993} and specifically in the context of BO \citep{locatelli_adaptive_1995,locatelli_bayesian_1997,osborne_gaussian_2009}. \citet{benassi_robust_2011} demonstrated that fully Bayesian optimisation (FBO) was statistically more robust in adversarial circumstances than ML-II. However, the study imposed strong assumptions, namely, the use of conjugate prior for output-scale, the lack of automatic relevance determination, and discretisation of length-scale. In today's hardware and software environments, these assumptions are unnecessarily restrictive and should be relaxed to obtain more general insights.

Over the past decade, a handful of studies adopted FBO that took advantage of more sophisticated algorithms and modern software platforms \citep{snoek_practical_2012,hernandez-lobato_predictive_2014,eriksson_high-dimensional_2021}. However, in these studies, FBO was adopted only because it served for evaluating their theoretical claims, and no further justification for the adoption was provided \citep{de_ath_how_2021}. Apart from these studies, the research on FBO remains sparse, and direct comparison between ML-II and FBO is virtually non-existent in the recent literature.

This paper provides three compelling reasons why researchers should consider adopting FBO for their applications where evaluations are expensive. First, failures of ML-II are more commonplace than implied by the existing studies which used the contrived scenarios. To support this argument, the experiments are set up only using common specifications, which other researchers may find in their applications, including those of the objective function, random initialisation, models, and algorithms. Second, FBO is more robust than ML-II, and the price of the robustness is almost trivial. This is shown in the results obtained from the identical experimental settings. In addition, to make the results relevant in modern applications, the restrictive assumptions in FBO specifications previously used in the literature are relaxed. Third, thanks to the modern software platforms, FBO is now simple to implement and fast enough to be practical. To demonstrate this, the experiments are implemented using Pyro \citep{bingham_pyro_2019} and BoTorch \citep{balandat_botorch_2020}. In addition, the code used to implement the experiments is written deliberately in a simple fashion so that it can serve as a template for other applications. The last point is a small but practical contribution because the existing code examples for FBO found in BoTorch package are written specifically for the particular study \cite{eriksson_high-dimensional_2021}, and modification for other FBO applications is not very straightforward.

In what follows, first, BO is reviewed as a technique to address a dynamic problem, which is deemed a more accurate formulation when involving evaluation budgets. The reader is reminded that, in this formulation, an optimal solution is intractable and suboptimality of BO is by design in practical applications (Section \ref{sec:BO}). Next, FBO is reviewed where the marginalised acquisition function and its approximation are highlighted as the key difference (Section \ref{sec:FBO}). Then, the experiments are described with emphasis of the roles played by Pyro and BoTorch, and the results are discussed (Section \ref{sec:experiments}). Finally, some concluding remarks are provided for future applications (Section \ref{sec:conclusion}). All the code and data are available on the website (\url{https://github.com/ysaikai/case4fbo}).

%%%%%%%%%%%%%%%%%%%%%%%%%%%%%%
\section{Bayesian optimisation}\label{sec:BO}
%%%%%%%%%%%%%%%%%%%%%%%%%%%%%%
A common formulation of the BO problem is a static optimisation problem:
\begin{equation}
    \min_{x\in\mathcal{X}} f(x)
\end{equation}
where \(f\) is the black-box objective function, and \(\mathcal{X}\) is a compact subset of Euclidean space. When BO is adopted in practice, however, it is often the case that function evaluations are expensive, and there is an evaluation budget. If an evaluation budget is explicitly modelled, a more accurate problem formulation is the following.

Let \(N\) be the total number of evaluations allowed. Then, the BO problem is a sequential decision problem; that is, choose \(x_n\in\mathcal{X}\) for each \(n\in\{1,\dots,N\}\) based on the history, the preceding observations \[\{(x_0,f(x_0)),\dots,(x_{n-1},f(x_{n-1}))\},\]
in order to minimise the best observed value after \(N\) evaluations:
\[\min_{n\in\{1,\dots,N\}} f(x_n).\]
Note that this is a dynamic problem as each choice depends on the preceding choices.

Since evaluations are allowed to be made only \(N\) times, it is necessary to alternatively search for an optimal sequence \(x_1^*,\dots,x_N^*\) without directly evaluating \(f\). This is where the notion of surrogate comes in. When BO with a GP surrogate is used to address the problem, an implicit assumption is that \(f\) is a sample path of the specified GP as a stochastic process. For the fixed GP, it is known that an optimal sequence \(\hat{x}_1,\dots,\hat{x}_N\) in terms of the Bayes risk is, in principle, obtained by dynamic programming \citep{betro_bayesian_1991}. However, except for trivial cases, the problem is numerically intractable \citep{benassi_robust_2011}. Furthermore, as mentioned above, specifying a suitable GP for a particular problem beforehand is rarely possible in practical problems. Instead, it is specified based on observations and sequentially updated throughout the trials.

The expected improvement (EI) acquisition function is a practical one-step lookahead strategy \citep{mockus_application_1978, jones_efficient_1998}. As a myopic strategy, EI is necessarily suboptimal \citep{ginsbourger_towards_2010} but proved reasonable in many applications \citep{shahriari_taking_2015}. Although many different acquisition functions have been proposed, EI maintains the practical advantage owing to the intuitive notion of ``improvement'', the relatively light computational requirement, and the proven performance.

%%%%%%%%%%%%%%%%%%%%%%%%%%%%%%
\section{Fully Bayesian optimisation}\label{sec:FBO}
%%%%%%%%%%%%%%%%%%%%%%%%%%%%%%
Performance of BO is dictated by the GP surrogate and the acquisition function. Since the former seems to take precedence over the latter \cite{shahriari_taking_2015}, and a GP is characterised by kernel hyperparameters, it is vitally important to carefully handle kernel hyperparameters and their uncertainty for successful BO applications. The principled way to handle parameter uncertainty is to follow the Bayesian formalism \citep{gelman_bayesian_2013}, according to which functions of uncertain parameters are weighted by the posterior probabilities of the parameters. (While the maximum a posteriori estimation is also used in practice to incorporate prior information, in the light of the Bayesian formalism, it does not fully reflect the parameter uncertainty and omitted in this paper.)

Let \(\theta\) and \(D_n\) denote respectively the hyperparameter and the data observed for the first \(n\) steps. Also, let \(\alpha(x|\theta,D_n)\) and \(p(\theta|D_n)\) denote respectively the acquisition function and the posterior of \(\theta\). Then, the marginalised acquisition function, the weighted average according to \(p(\theta|D_n)\), is
\begin{equation}\label{eq:exact}
    \alpha(x|D_n) = \mathbb{E}_{\theta|D_n}\alpha(x|\theta,D_n) = \int \alpha(x|\theta,D_n)p(\theta|D_n)d\theta.
\end{equation}
In FBO, this is the function to maximise for the next query point
\begin{equation}
    x_{n+1} \in \argmax_{x\in\mathcal{X}
    }\alpha(x|D_n).
\end{equation}
In other words, due to the randomness in \(\theta\), it is a stochastic optimisation problem. Since the above integral generally permits no analytic expression for \(\alpha(x|D_n)\), in practice, use of approximation methods is required to carry out the maximisation. In BO contexts, the maximisation is commonly addressed by the sample average approximation \citep{homem-de-mello_monte_2014}, which is well supported by BoTorch as reviewed below.

Except for trivial cases, the posterior \(p(\theta|D_n)\) is a complicated distribution, for which no off-the-shelf sampler is available, so it is necessary to use some approximation methods to obtain samples. Among many different methods, Hamiltonian Monte Carlo is the gold standard \citep{eriksson_high-dimensional_2021} and, in particular, the No-U-Turn sampler \citep{hoffman_no-u-turn_2014} is an attractive option as it spares researchers the trouble of manually tuning the step size and path length in Hamiltonian Monte Carlo.

After obtaining \(M\) Monte Carlo samples, the marginalised acquisition function is approximated by the finite sum:
\begin{equation}\label{eq:approx}
    \alpha(x|D_n) \approx \frac{1}{M}\sum_{m=1}^M \alpha(x|\theta_m,D_n).
\end{equation}
Given \(M\) samples, the right-hand side is a deterministic function, which is commonly maximised using quasi-Newton methods such as L-BFGS-B \cite{byrd_limited_1995}.

%%%%%%%%%%%%%%%%%%%%%%%%%%%%%%
\section{Experiments}\label{sec:experiments}
%%%%%%%%%%%%%%%%%%%%%%%%%%%%%%
\subsection{The test problem}
To compare performances of the two approaches, ML-II and FBO, the experiments were conducted using common specifications, which many researchers may find in their applications. The performance metrics is simple ``regret'' after \(N\) evaluations, difference between the true minimum and the best observed value over \(N\) evaluations. Different \(N\in\{1,\dots,30\}\) corresponds to a different evaluation budget. The following table contains key specifications of the test problem.

\begin{table}[H]
\begin{center}
\begin{tabular}{ l l }
  Objective function & Ackley function on \(\mathcal{X}\subset\mathbb{R}^2\)\\ \hline
  GP Kernel & Matérn 5/2\\ \hline
  GP Mean function & 0 for all \(x\)\\ \hline
  Noise variance & \(10^{-6}\)\\ \hline
  Prior for output-scale & Log normal with mean 10 and standard deviation 10\\ \hline
  Prior for length-scale & Log normal with mean 0.5 and standard deviation 0.5
\end{tabular}
\end{center}
\end{table}

The objective function was Ackley function \citep{ackley_connectionist_1987}, one of the common optimisation test functions \citep{surjanovic_optimization_nodate}. The function was specified by following the recommendation \citep{surjanovic_optimization_nodate}, which is also implemented and the default setting in BoTorch. For example, the function domain was set to \(\mathcal{X}=[-32.768, 32.768]\times[-32.768, 32.768]\).

The total 101 experiments were conducted, corresponding to 101 distinct \(x_0\) and random seeds (0-100) appropriately set for reproducibility. Fixing random seeds was necessary not only for the hyperparameter sampling but also for the random multistart used by BoTorch in optimisation of acquisition functions and marginal likelihood. Each experiment was initialised by \(x_0\), which was shared between ML-II and FBO. Then, the subsequent \(N\) observations were made by following each approach.

The acquisition function is EI, which works relatively well for noise-less test functions \citep{de_ath_how_2021}. The class of GP is a basic one---the zero-mean function, Matérn(5/2) kernel with automatic relevance determination, and an output-scale parameter. Although the objective function is noise-less, following the convention, the noise variance is set to \(10^{-6}\) for numerical stability. As a result, there were three hyperparameters: one output-scale and two length-scales. In FBO, three prior distributions were assumed to be independent.

While gamma distribution is also common, for both output-scale and length-scales, log normal distribution was adopted as it has strictly diminishing probability towards 0, which helps to avoid sampling unreasonably small length-scales, which could in turn cause numerical problems. The mean for each prior was chosen based on empirical evidence, while the standard deviation was set so that the relative standard deviation was equal to 1.

Note that the prior for length-scale reflects the fact that inputs to GP are normalised to between 0 and 1. Note also that output observations used for GP are standardised to have mean 0 and standard deviation 1, which justifies the use of the zero-mean function. The sampler was Hamiltonian Monte Carlo and the No-U-Turn sampler. Following the recommendation \citep{eriksson_high-dimensional_2021}, at each iteration, \(512\) samples were first drawn as warm-up/burn-in samples and then \(256\) samples were drawn, which were thinned to \(M = 16 = 256/16\) samples by keeping only every 16th sample.

Many of the existing software implementations specify log normal distribution with two parameters for the mean and standard deviation of the corresponding normal distribution. In this case, when the desired mean and standard deviation of log normal distribution are \(\mu\) and \(\sigma\), following are the corresponding parameters of normal distribution:
\begin{align*}
    \mu_N &= \log\mu - \frac{1}{2}\log\left(\frac{\sigma^2}{\mu^2}+1\right)\\
    \sigma_N &= \sqrt{\log\left(\frac{\sigma^2}{\mu^2}+1\right)}.
\end{align*}

\subsection{Software platforms}
Software platforms play a crucial role in modern machine learning research and practice. In BO applications, key numerical operations include computation of posterior GP, optimisation of acquisition functions, optimisation of marginal likelihood (for ML-II), and sampling of hyperparameters (for FBO). In this paper, Pyro and BoTorch were used to implement the experiments. The software versions used for the experimentation are 3.9.7, 1.8.0, and 0.6.4 for Python, Pyro, and BoTorch respectively.

Pyro facilitates posterior sampling of kernel hyperparameters. In general, Hamiltonian Monte Carlo and the No-U-Turn sampler are sophisticated algorithms that compute the gradient of the posterior probability density and solve the ordinary differential equations based on the Hamiltonian while automatically tuning the step size and path length for the solver. As a result, manual implementation requires a significant amount of coding. With Pyro, however, the implementation becomes quite simple and straightforward; essentially, it is only necessary to specify the priors and desired number of samples.

BoTorch facilitates implementation of GPs and acquisition functions, computation of posterior GP, optimisation of marginal likelihood, and optimisation of acquisition functions. While in FBO optimisation of the acquisition function takes place using the sample average approximation (Eq.\ref{eq:approx}), BoTorch is designed to handle this operation efficiently.

Both platforms are built upon PyTorch and in particular take advantage of its automatic differentiation functionality, which facilitates computing the gradient in optimisation by BoTorch and Hamiltonian Monte Carlo by Pyro. Moreover, both platforms automatically implement parallel processing and utilise multiple cores of CPU/GPU if available.

\subsection{Results and Discussion}
\begin{figure}[H]
    \centering
    \includegraphics[width=\linewidth]{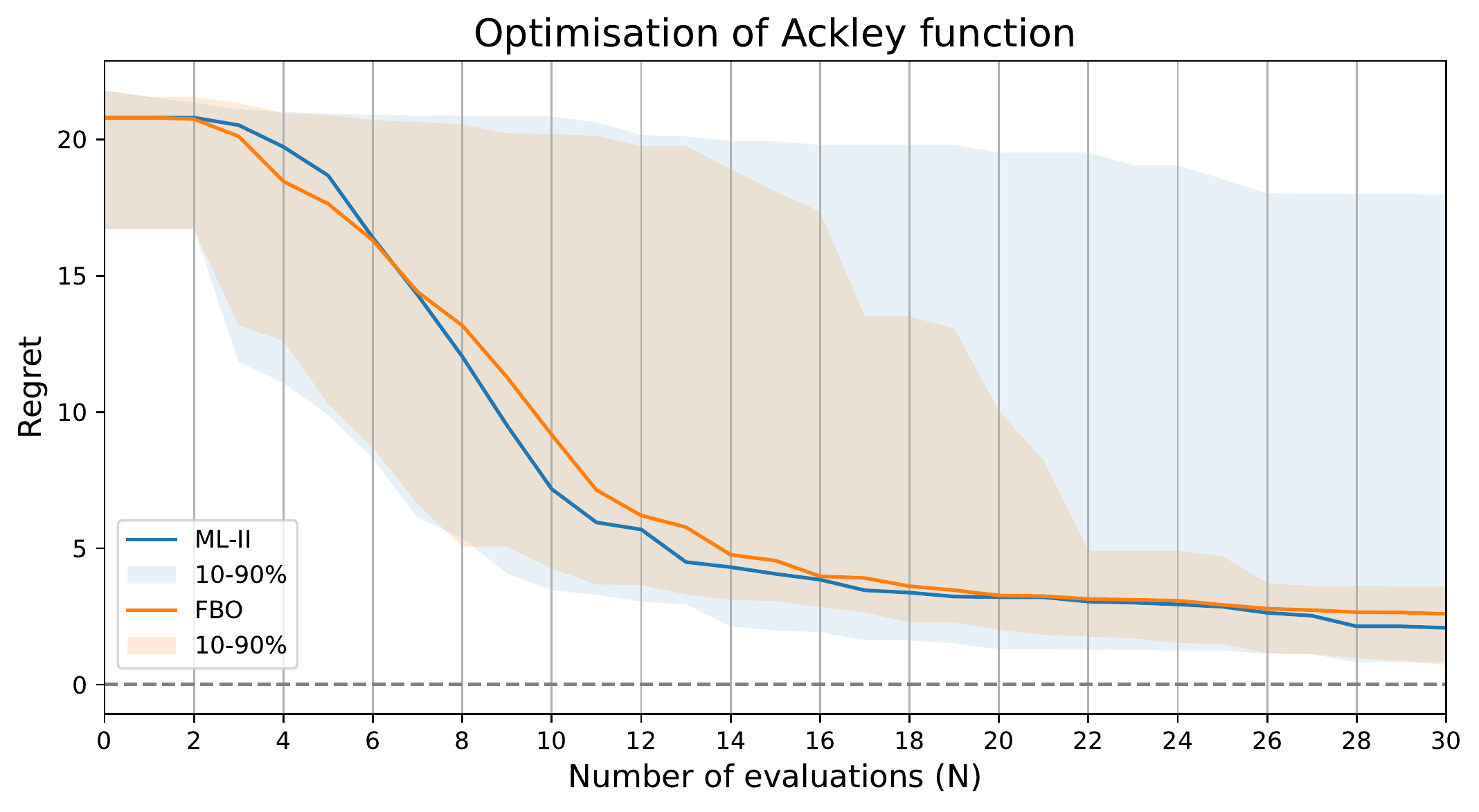}
    \caption{The results from 101 experiments for each number of evaluation budgets (N=1,...,30). 101 experiments are distinguished from each other by 101 distinct initial samples \(x_0\) and random seeds. The solid lines represent the median values. The shaded regions are formed by the 10th and 90th percentiles.}
    \label{fig:results}
\end{figure}

\begin{figure}[H]
    \centering
    \includegraphics[width=\linewidth]{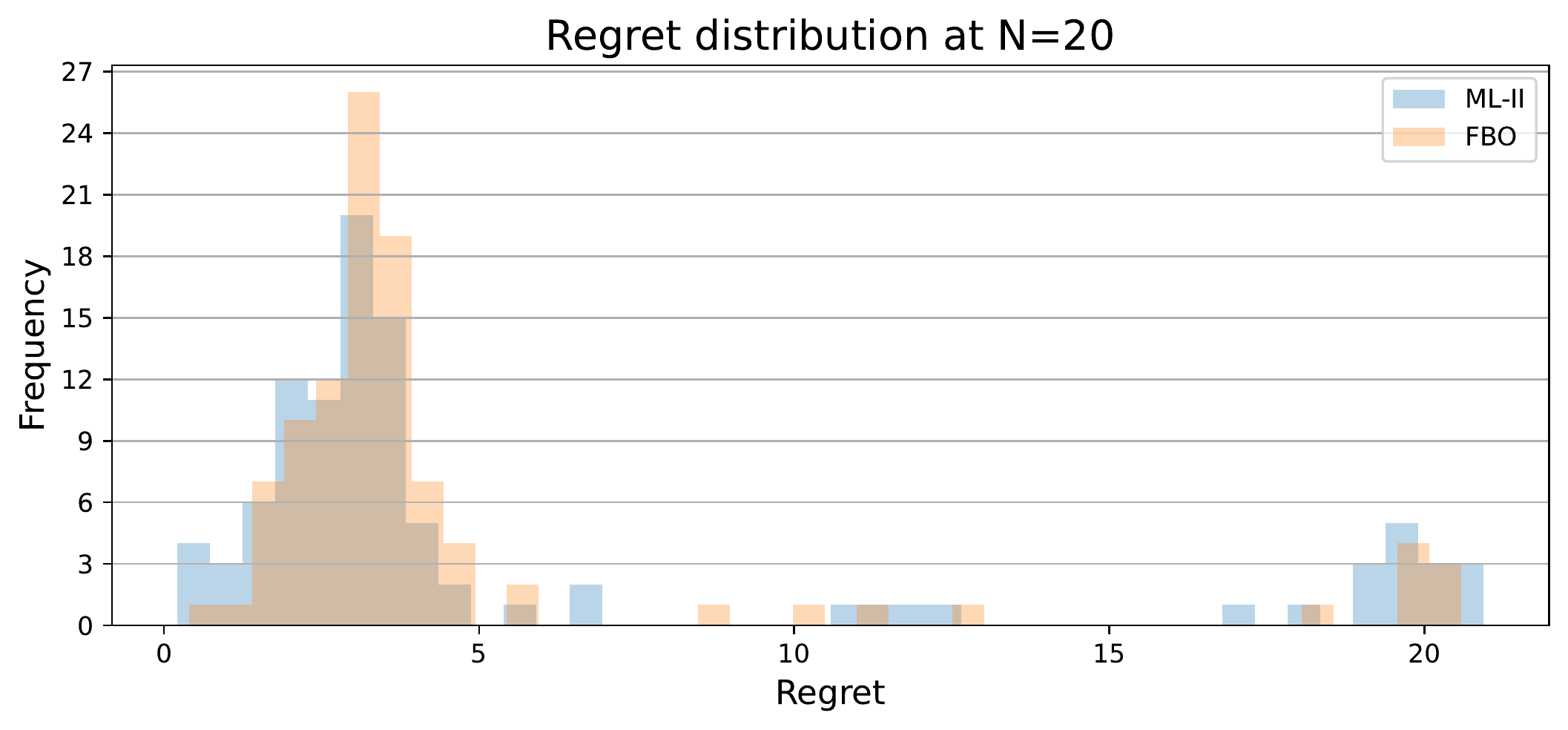}
    \caption{Regret distribution of 101 samples after 20 evaluations.}
    \label{fig:hist20}
\end{figure}

\begin{figure}[H]
    \centering
    \includegraphics[width=\linewidth]{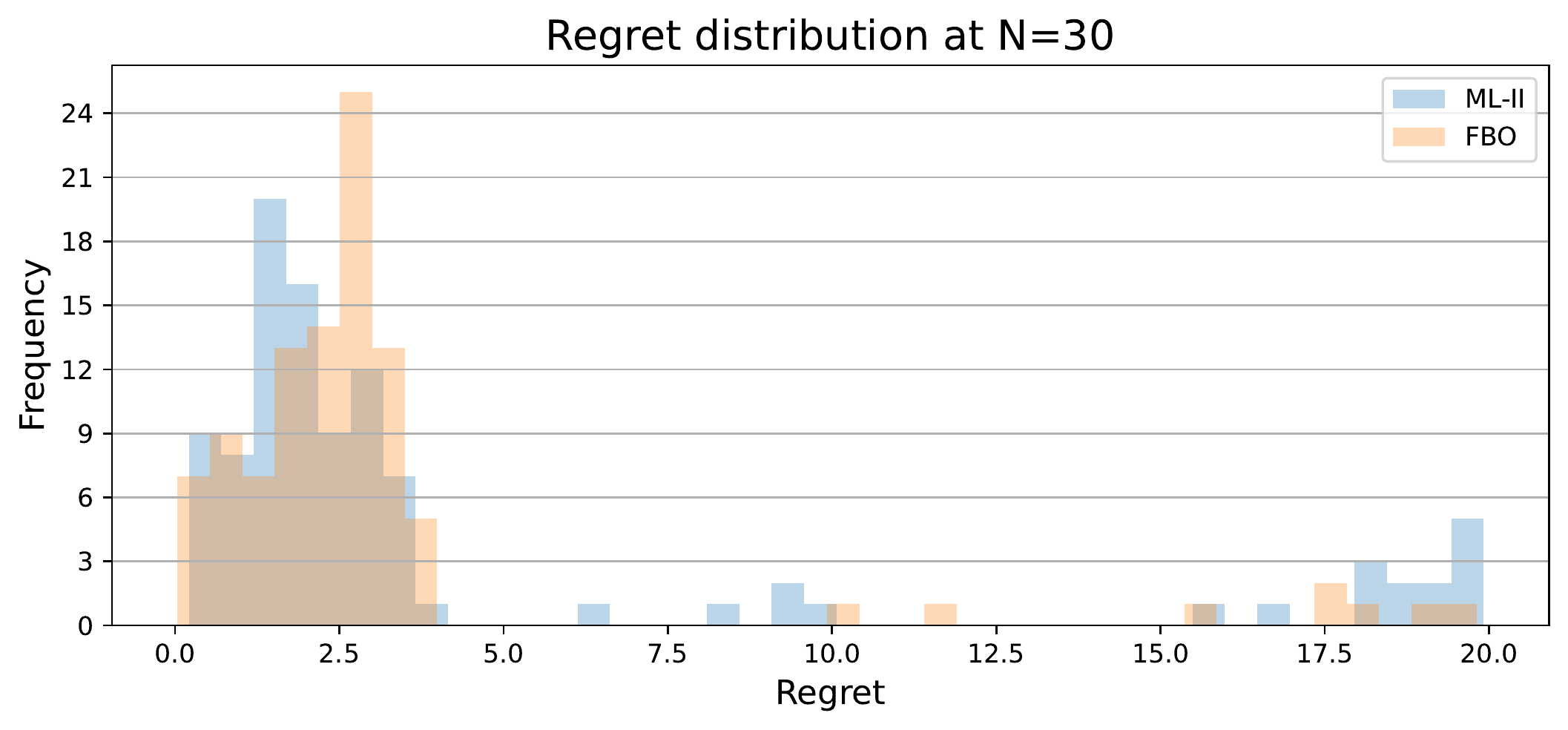}
    \caption{Regret distribution of 101 samples after 30 evaluations.}
    \label{fig:hist30}
\end{figure}

Figure \ref{fig:results} illustrates the results by plotting the regret at each evaluation budget \(N\in\{1,\dots,30\}\) for both ML-II and FBO in terms of median and region formed by the 10th and 90th percentiles. The regret distribution at \(N=0\), which is identical between ML-II and FBO, reflects the distribution of function evaluations \(f(x_0)\) at 101 random initial \(x_0\). This initial distribution largely remains unchanged at \(N=1\) and \(N=2\). Then, indicated by the downward-sloped median curves, performances of both approaches start to improve at \(N=3\) and continue until \(N=30\), although the rate of improvement slows down after \(N\approx10\). Throughout all the evaluation budgets, the median performances and downward dispersion from the medians are very similar between ML-II and FBO. In contrast, as \(N\) increases, two distributions in terms of upward dispersion increasingly diverge. While the upward dispersion for FBO continues to shrink, the one for ML-II remains large, indicating the greater number of failures. For more details, Figure \ref{fig:hist20} and \ref{fig:hist30} plot histograms for \(N=20\) and \(30\) respectively.

Given the strong similarity in median and downward dispersion, the shrinking upward dispersion indicates that FBO has clear advantage over ML-II in terms of statistical robustness. However, bear in mind that the advantage is only statistical, and FBO certainly can fail as indicated by the occurrence of large regrets in Figure \ref{fig:hist20} and \ref{fig:hist30}. First of all, since FBO uses sampling of random hyperparameters, poor choices of output-scale and length-scale can happen. However, the influence of these unlucky samples can be reduced by increasing the number of Monte Carlo samples (\(M\) in Eq.\ref{eq:approx}).

More fundamentally, even if using the exact marginalised acquisition function (Eq.\ref{eq:exact}), FBO with EI (or any other practical acquisition function) is still myopic and may fail when \(x_0\) is unfavourable. As analysed in the literature, such unfavourable initial samples ``fool'' the interpretation of data and lead to ``deceptive'' GPs with ill-updated hyperparameters \citep{forrester_global_2008, benassi_robust_2011}, which in turn result in uninformative acquisition functions and undermine the effectiveness of BO. In real-world applications where function evaluations are expensive, researchers may not have enough information about problems and have to choose initial samples more of less randomly. Consequently, the failures caused by random \(x_0\) in the experiments are deemed a relevant warning to researchers.

What FBO can achieve is to dilute the harm by adopting a continuum of data interpretations and averaging them out according to the posterior distribution of hyperparameters. Even 10 years ago, without restrictive assumptions, FBO might not have been a practical option. Now, thanks to the recent development of both software and hardware technologies, it can be a viable option for many practical problems. Given the ease of implementation, the price of the robustness advantage of FBO is only additional computation, which is approximately 10 times greater than for ML-II. Considering the expensiveness of experiments, however, the extra computing time is relatively trivial in many applications. Thus, use of FBO is a simple way to guard against potential failures that end up wasting precious research resources.

%%%%%%%%%%%%%%%%%%%%%%%%%%%%%%
\section{Conclusion}\label{sec:conclusion}
%%%%%%%%%%%%%%%%%%%%%%%%%%%%%%
While small-sample performance is the main reason to use Bayesian optimisation when function evaluations are expensive, the standard approach based on type II maximum likelihood may fail in many practical applications. The paper made the case for fully Bayesian optimisation as a more robust and viable alternative. Thanks to the modern software and hardware technologies, the implementation is simple, and the execution is fast enough to be practical in many cases. Since the benefits seem to outweigh the costs, researchers should consider adopting fully Bayesian optimisation for their applications so that they can guard against potential failures that end up wasting precious research resources.

%%%%%%%%%%%%%%%%%%%%%%%%%%%%%%
\printbibliography
%%%%%%%%%%%%%%%%%%%%%%%%%%%%%%

\end{document}